\documentclass[sigconf,nonacm]{acmart}

\AtBeginDocument{%
  \providecommand\BibTeX{{%
    \normalfont B\kern-0.5em{\scshape i\kern-0.25em b}\kern-0.8em\TeX}}}

\usepackage[symbol]{footmisc}

\begin{document}

\title{FlexiFilm: Long Video Generation with Flexible Conditions}

%
\author{Yichen Ouyang}
\email{22271110@zju.edu.cn}
\affiliation{%
  \country{Zhejiang University}
}

\author{Jianhao Yuan}
\email{yuanjianhao2019@gmail.com}
\affiliation{%
  \country{Oxford University}
}

\author{Hao Zhao}
\email{zhaohao@air.tsinghua.edu.cn}
\affiliation{%
  \country{Tsinghua University}
}

\author{Tiejun Huang}
\email{tjhuang@pku.edu.cn}
\affiliation{%
  \country{BAAI and Peking University}
}

\author{Gaoang Wang}
\thanks{\dag Corresponding authors}
\authornotemark[2]
\email{gaoangwang@intl.zju.edu.cn}
\affiliation{%
  \country{Zhejiang University}
}

\author{Bo Zhao}
\authornotemark[2]
\email{bozhao@pku.edu.cn}
\affiliation{%
  \country{BAAI}
}



\begin{abstract}
Generating long and consistent videos has emerged as a significant yet challenging problem. While most existing diffusion-based video generation models, derived from image generation models, demonstrate promising performance in generating short videos, their simple conditioning mechanism and sampling strategy—originally designed for image generation—cause severe performance degradation when adapted to long video generation. This results in prominent temporal inconsistency and overexposure. Thus, in this work, we introduce FlexiFilm, a new diffusion model tailored for long video generation. Our framework incorporates a temporal conditioner to establish a more consistent relationship between generation and multi-modal conditions, and a resampling strategy to tackle overexposure. Empirical results demonstrate FlexiFilm generates long and consistent videos, each over 30 seconds in length, outperforming competitors in qualitative and quantitative analyses. Project page: \url{https://y-ichen.github.io/FlexiFilm-Page/}.

\end{abstract}

\maketitle

\section{Introduction}
\label{sec:intro}

Recently, there has been a significant progress in video generation models~\cite{blattmann_stable_nodate,hong_cogvideo_2022,kondratyuk_videopoet_2023, wu_tune--video_2023, villegas_phenaki_2022}, revolutionizing media synthesis technology and providing a variety of possibilities for artistic creation and data augmentations.
Most of the existing methods employ diffusion models~\cite{chen_videocrafter1_2023, blattmann_stable_nodate, he_latent_2022} with additional temporal adapter inserted. For example, Stable Video Diffusion~\cite{blattmann_stable_nodate} initiates the model weights from a pretrained image diffusion model with trainable temporal layer, and generates high-fidelity frames after training on video datasets. 

\begin{figure}[t]
  \centering
  \includegraphics[width=.5\textwidth]{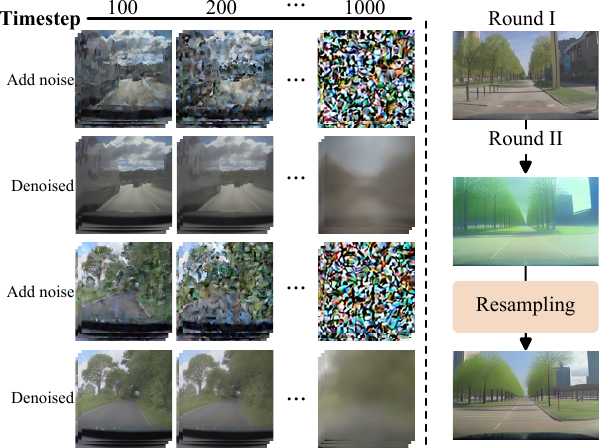}
  \caption{Non-zero SNR problem. In the left subplot, the last columns of rows 2 and 4 are the denoised outputs of model from two noisy images at t=1000, but they have different mean distributions. In the right subplot, after 2 round of inference (using the last frame generated on round 1 as the condition for round 2), the generated frames become overexposed, but this can be solved with our resampling strategy.}
  \label{fig:snr}
\end{figure}

While these diffusion-based video generation model demonstrates promising capacity in generating high-fidelity content, the generated videos are usually restricted to a few seconds with dramatic performance degradation when adopting long video generation tasks. 
In particular, we observe two main failure modes: \textit{temporal inconsistency} and \textit{overexposure}. 
The former arises from an insufficient conditioning mechanism, as current video diffusion models ~\cite{zhou_magicvideo_2022,hong_cogvideo_2022,blattmann_stable_nodate} typically utilize a text description or a single image as the conditioning input, which contains insufficient temporal information to handle the evolving dynamics and complex semantic in long video generation. 
The latter issue stems from a deficient sampling strategy, wherein the noise scheduler fails to enforce a zero Signal-to-Noise Ratio (SNR) at the conclusion of the denoising step, resulting in what is known as the non-zero SNR problem (as described in Sec~\ref{sec:related-SNR} and Fig~\ref{fig:snr}). Such miscalibration leads to overexposure or structural collapse, particularly when generating longer videos through multiple rounds of inference in a recursive manner.

To address these challenges, we propose FlexiFilm, a latent generative model for more controllable and high-quality long video generation. The framework consists with two main component of a temporal conditioner and a resampling strategy, along with a co-training method.
To address the problem of insufficient conditioning, we propose a temporal conditioner to flexibly extend the conditions and provide guidance to videos with any number of generated frames, providing rich information to help the model learn the temporal relationship between the generating frames and the conditional frames. In addition, we propose a co-training method for temporally consistent condition-generation module collaboration.
To address the problem of overexposure resultant from multiple round inference, we introduce a resampling strategy in the inference stage to solve the non-zero SNR problem, which helps the model to maintain the original quality when generating much longer videos in a recursive manner. 
Extensive experiments validate the effectiveness of our proposed framework in generating longer and more consistent videos than the baseline methods in both visual quality and evaluation metrics as shown in Fig~\ref{fig:exp}.
The primary contributions of our work are as follows:
\begin{itemize}
    \item A temporal conditioner to guide video diffusion models with more modals with richer information, modeling long-term relationship between generation and conditions.
    \item A co-training method that greatly improves the inter-frame consistency of the generated video, by collaboratively training the temporal modules in the conditioner and U-Net.
    \item A resampling strategy for multi-round inference to alleviate non-zero SNR problems, enabling much longer video generation in a recursive manner without harming the quality.
\end{itemize}

\begin{figure*}[t!]
  \centering
  \includegraphics[width=\textwidth]{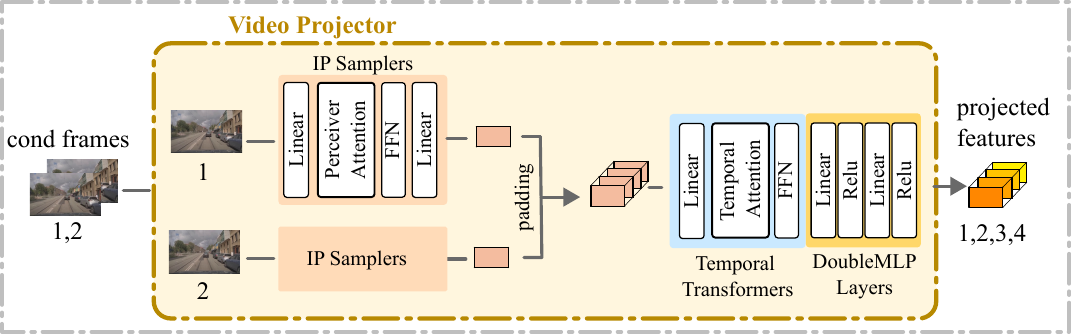}
  \caption{Structure of video projector. In the proposed video projector, the condition frames pass through IP samplers separately to obtain independent spatial information, and then go through temporal transformers together to learn inter-frame temporal information. After that, the finally obtained projected feature contains rich information both spatially and temporally.  }
  \label{fig:model}
\end{figure*}

\section{Related Works}
Video generation has been a compelling yet challenging area of research. One line of emerging works is derived from the general family of transformer-based language models ~\cite{wang_lavie_2023,kondratyuk_videopoet_2023,hu_gaia-1_2023,yu_language_2023}, where they leverage visual tokenizers and text tokenizers to obtain the discrete token of each video frame. Subsequently, they train a decoder-only transformer to learn to generate subsequent frames autoregressively. Although this autoregressive framework is naturally suitable for long video generation, the prohibitively expensive computational requirements of large-scale training data is often unrealistic for common researchers ~\cite{kondratyuk_videopoet_2023}. Thus, in this work, we focus on an alternative approach to video generation that relies on diffusion models to generate multiple frames with text or images conditions ~\cite{zhang_show-1_2023, blattmann_stable_nodate, luo_videofusion_2023, shi2024motion, henschel2024streamingt2v}. These video diffusion models are often derived from pretrained text-to-image (T2I) diffusion models by incorporating additional adapter of various forms. For example, VideoCrafter ~\cite{chen_videocrafter1_2023} adds temporal layers to 3D U-Net diffusion models and initializes the spatial layers' weights from Stable Diffusion models ~\cite{rombach2022highresolution}. AnimateDiff ~\cite{guo2023animatediff} incorporates additional motion module in pre-trained T2I diffusion model to explicitly learn and encode motion prior guiding video generation. While these diffusion-based models exhibit a strong capacity for generating high-fidelity visual content, they are generally capable of generating only short videos of a few seconds. This limitation is due to their simple conditional mechanism and sampling strategy, which were originally designed for image generation. Specifically, the insufficient conditioning fails to provide the necessary temporal information required to manage the evolving dynamics and complex semantics in long video generation, leading to temporal inconsistencies. Additionally, the deficient sampling strategy, when applied under recursive inference, introduces a miscalibrated signal-to-noise ratio, resulting in overexposure or structural collapse in the generated frames.

Various solutions have been proposed to achieve long video generation. Several studies ~\cite{yin2023nuwa,zhou_magicvideo_2022, chen2023seine} adopt a two-stage generation method: first, they generate keyframes guided by the input reference, and then interpolate these keyframes to obtain longer videos. Other works aim to improve the sampling strategy to enhance consistency. For instance, FreeNoise ~\cite{qiu2024freenoise} reschedules a sequence of noises for long-range correlation and performs temporal attention to improve consistency in long videos. Additionally, other research ~\cite{wang2023gen} treats long video as a collection of short clips and generates long videos through a parallel set of inferences with controlled temporal overlapping. Although these methods alleviate the problem of generating longer videos, they struggle to balance both video quality and consistency.

In this work, we follow the U-Net architecture of video diffusion models ~\cite{blattmann_align_2023,he_latent_2022}, and design a flexible temporal conditioner to support the generation of videos that maintain both high quality and long-term consistency. This temporal conditioner can process a single image or multi-frame videos, providing guidance with rich temporal and spatial information for the long video generation process.

\section{Method}
\label{sec:method}

\subsection{Overview}
In this section, we introduce the FlexiFilm framework for synthesizing long-and-consistent videos. We first introduce the model architecture in Sec~\ref{sec:method-model}, along with a novel temporal conditioner for multi-modal guidance, modeling temporal relationship between long-term generations and conditions. Then we introduce the temporal module co-training strategy in Sec~\ref{sec:method-train}, designing for inter-frame consistent video generation. After that, we introduce a resampling strategy in Sec~\ref{sec:method-infer} for alleviating non-zero SNR problem in multi-round generation and enabling much long video generation. Finally, we introduce our dataset FF-drive1 in Sec~\ref{sec:method-dataset} for long video generation task, in form of detailed video-text pairs.

\begin{figure*}[t!]
  \centering
  \includegraphics[width=\textwidth]{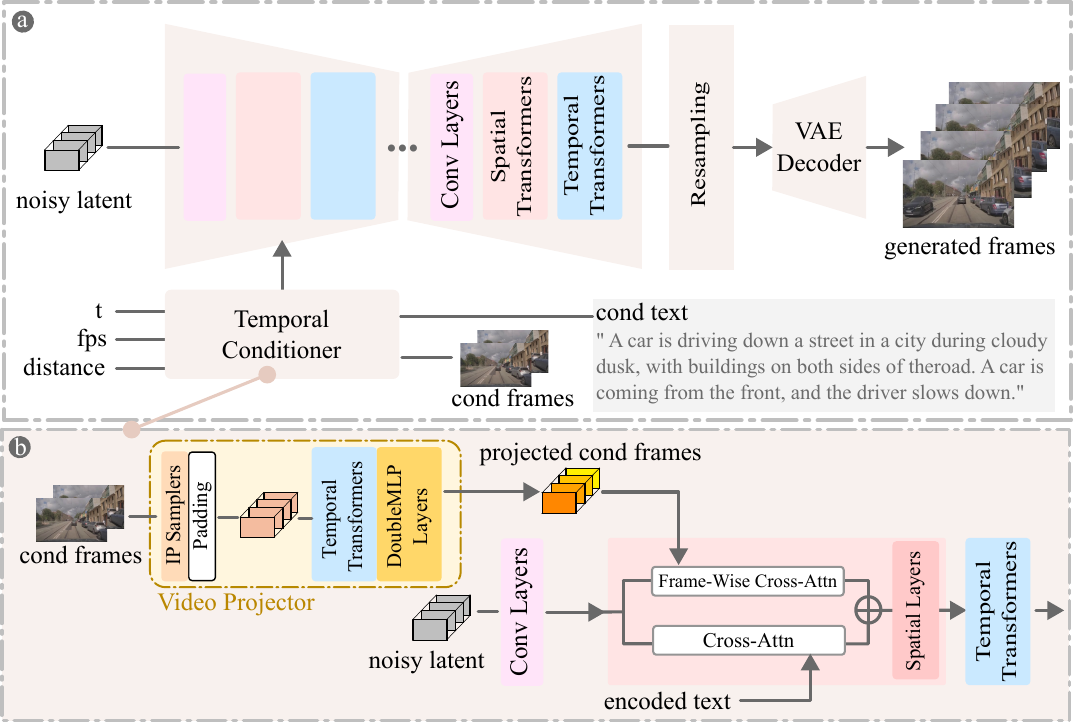}
  \caption{FlexiFilm pipeline. The subplot (a) shows the backbone of FlexiFilm is a 3D U-Net working on the latent space of VAE, using a temporal conditioner for multi-modal (text, image or video) referencing for video frames generate. The subplot (b) shows the workflow of the proposed temporal conditioner, where visual contents (image or frames) and text contents are fused to guide the video generation process with both spatial and temporal information.}
  \label{fig:pipeline}
\end{figure*}

\subsection{Model Architecture}
\label{sec:method-model}
The FlexiFilm model consists of a VAE, a 3D U-Net, and a temporal conditioner as Fig~\ref{fig:pipeline}(a) shows. Precisely, we adopt the pretrained VAE~\cite{kingma2022autoencoding} to the input video $x_0$, and project it frame by frame to video latent $z_0$. The 3D U-Net is initiated from a pretrained VideoCrafter-1 model ~\cite{chen_videocrafter1_2023}, which is a kind of latent video diffusion model ~\cite{he_latent_2022}. Then the diffusion process is performed on the distribution of $z_0$ after obtaining the video latents, with the training process discussed in Sec~\ref{sec:method-train} and the inference process discussed in Sec~\ref{sec:method-infer}. The temporal conditioner allows long text description and image or frames to be input as the multi-modal reference, of which the structure and workflow is shown in Fig~\ref{fig:pipeline}(b). 

Our insight for designing the temporal conditioner is to model temporal relationship between conditional frames and generating frames. As Fig~\ref{fig:pipeline}(b) shows, the conditional frames pass through IP samplers $VP_{S-Attn}$ and a temporal transformer $VP_{T-Attn}$ in the video projector to obtain information-rich features. And then the projected features  is sent to spatial transformers of 3D U-Net $UNet_{S-Attn}$ to do frame-wise cross attention with the proceeding noisy video latent $z_0$. And the pipeline can be described as:
\begin{equation}
\begin{split}
F_{generate}&:=\mathcal{D}(\hat{z}_{\epsilon, cond, r}),\\
cond:= ((VP_{S-Attn}(F_{cond}) \odot &VP_{T-Attn})+ \mathcal{E}(text))\odot UNet_{S-Attn}
\end{split}
\end{equation}
where the $F_{generate}$ is generating frames, $\mathcal{D}$ is the VAE Decoder, $\hat{z}$ is the denoising video latent, $\epsilon$ is a random Gaussian noise, $r$ represents for the resampling score, and $\mathcal{E}$ is the text encoder.

In the proposed temporal conditioner in model architecture, the core component is a novel video projector, as shown in Fig~\ref{fig:model}.
Take the workflow of 2-frame condition and 4-frame generation as an example. In the video projector, the 2-frame conditions first passes through IP samplers separately to obtain the features of each frame, which only contain the spatial information of each frame independently. Then the features of the last frame are padded to 4-frame and sent together to a temporal transformer to obtain relationship between conditional frames, and finally output projected 4-frame condition feature through a double-mlp layers, which then performs frame-wise attention with the 4-frame generating features.

\subsection{Training Strategy}

\label{sec:method-train}
\subsubsection{Scalable Overlapping Design}
During the training process, the 3D U-Net learns to gradually denoise the noisy target under condition frames. It is notable that the conditional and the generative frames overlap at the first $n_c$ frames. Our insight for designing the condition and target is that the overlapping makes training easier and converging faster. Besides, we believe that there is also a scalable relationship in the video generation process: the more overlap frames, the closer the distribution of the generated video and the condition frames will be. Therefore, we design a variable number of overlap video frames for training and adding change factors to the condition, and this allows the model to better understand video generating process, enabling better quality for image-to-video and frames-to-video generation.

\begin{figure}[h]
  \centering
  \includegraphics[width=.5\textwidth]{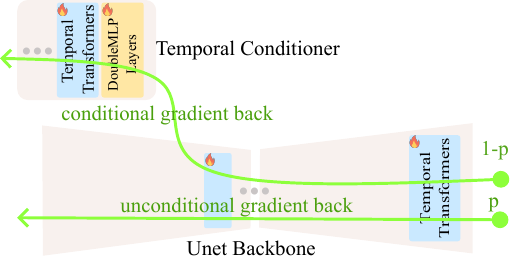}
  \caption{Temporal module co-training. All modules containing learnable parameters are drawn with a fire icon, and the two paths of gradient pass-backing are marked with green arrows, occurring with probabilities $p$ and $p-1$ respectively.}
  \label{fig:train}
\end{figure}

\subsubsection{Temporal-Consist Co-Training}
In addition, we model the temporal relationship between the conditional frame $F_{cond}$ and the generated latent $z_{generate}$ by co-training the temporal conditioner's temporal modules $VP_{T-Attn}$ and the U-Net's temporal module $UNet_{T-Attn}$, while freezing other parts of the model, and the trainable parameters $\theta,\varphi$ are defined in \ref{fml:train}:
\begin{equation}
Trainable(\theta,\varphi) := VP_{T-Attn}^\theta(F_{cond}) \odot UNet_{T-Attn}^\varphi(z_{generate})
\label{fml:train}
\end{equation}
During the co-training, we use a mixed training strategy of conditional model and unconditional model with probabilities as shown in Fig~\ref{fig:train}. Specifically, we set the condition of the model to empty with a certain probability $p$. Therefore, the unconditional model is trained at probability at $p$, and the conditional model at $1-p$, It is notable that, for the unconditional model, only the learnable parameters in the U-Net backbone will be trained, while for the conditional model both the learnable parameters in temporal conditioner and U-Net backbone will be trained. In experiment, we find that the model performs best when $p= 0.1$, which is intuitively reasonable: the framework trains the unconditional model with a probability of $0.1$, and trains the conditional model with a probability of $0.9$, which means that the conditioner got adequate training, and since the U-Net backbone has been initialized from the pretrained checkpoints, a small degree of fine-tuning is enough.

\subsection{Inference Strategy}
\label{sec:method-infer}
\subsubsection{Multi-Round Inference for Longer Video Generation}
During the inference stage, we first use the DDIM sampler to sample a video $v$ from a pure Gaussian noise $\epsilon$ with classifier-free-guidance (CFG)~\cite{ho2022classifierfree} scale at $s$, where the condition is text and image/frames. Considering that by 1-round inference, the sampled video $v$ has $f$ frames, then we can use the last few frames of $v$ as the condition, and do a second round of inference to get the following $f$ frames. Therefore, by recursively sampling $m$ rounds as this, the video diffusion models can finally generate $m*f$ frames theoretically.
In previous works, they often suffer from non-zero SNR problem~\cite{lin_common_nodate} that occurs during multi-round inference: the generated video suffers from over-exposure problem due to high CFG, and the generated frames quickly collapse with the increasing of round $m$. 

\subsubsection{Resampling Strategy for Non-Zero SNR Problem}
\label{sec:related-SNR}
The non-zero signal-to-noise (SNR) ratio problem is proposed by Lin et al. ~\cite{lin_common_nodate}: the common diffusion noise scheduler do not enforce the last timestep to have zero SNR but the model is given pure Gaussian noise at inference, thus creating a discrepancy between training and inference. During the inference stage, since the starting point of the inference process has an SNR close to zero, classifier-free guidance becomes very sensitive and cause images to be overexposed. In multiple rounds of inference, this problem will cause the model to accumulate errors, causing the multi-round generated frames to quickly distort and collapse as Fig~\ref{fig:ablation_eval} shows.
To solve this problem and enable multi-round inference without harming the video quality, we propose a resampling strategy as Eq~\ref{fml:resample} shows.
\begin{equation}
\begin{split}
z = z_{neg} + s \cdot (z_{pos}-&z_{neg}), \sigma_{pos} = std(z_{pos}), \sigma_{z} = std(z)\\
z_{resampled} &= r\cdot \frac{\sigma_{pos}}{\sigma_{z}} \cdot z + (1-r)\cdot z
\end{split}
\label{fml:resample}
\end{equation}

In Eq~\ref{fml:resample}, we rescale the latent $z$ of video $v$ back to the original standard deviation before applying CFG and introduce a resampling scale $r$ to control the rescaling strength.

\begin{table}[htb] 
\centering
\begin{tabular}{l|c}
\toprule
\textbf{Method} & \textbf{FVD$\downarrow$} \\
\midrule
InternVideo ~\cite{wang2022internvideo} & 617 \\
CogVideo (ZH) ~\cite{hong_cogvideo_2022} & 751 \\
Cogvideo (En) ~\cite{hong_cogvideo_2022} & 702 \\
MagicVideo ~\cite{zhou_magicvideo_2022} & 655 \\
Make-A-Video ~\cite{singer2022make} & 367 \\
Video LDM ~\cite{blattmann2023align} & 550 \\
PYOCO ~\cite{ge2023preserve} & 355 \\
Show-1~\cite{zhang_show-1_2023} & 394 \\
VideoPoet (Pretrain)~\cite{kondratyuk_videopoet_2023} & 355 \\
\midrule
FlexiFilm (Ours) & \textbf{336} \\
\bottomrule
\end{tabular}
\caption{UCF-101 zero-shot video generation comparison.}
\label{tab:fvd_scores}
\end{table}

\begin{figure*}[t!]
  \centering
  \includegraphics[width=\textwidth]{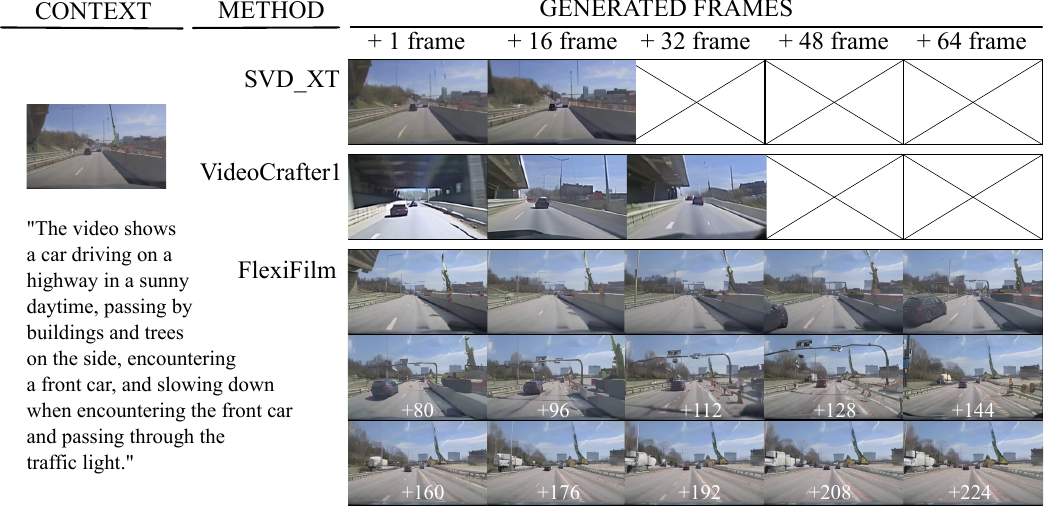}
  \caption{Comparison with baselines. The comparison results show the good performance of our model in long video generation (200+ frames), while the baselines produce less consistent results and can generate only short clips (32- frames). In the video generated by SVD, as the vehicle moves forward, the buildings unreasonably maintain their original size; while the video generated by VideoCrafter has a different structure from the reference image and lack inter-frame consistency.}
  \label{fig:exp}
\end{figure*}

\subsection{Dataset Construction}
\label{sec:method-dataset}
\subsubsection{Insight for the Dataset}
The goal of long video generation is to ensure the spatial and temporal consistency of the generated video over a longer period of time, so the generation process requires fairly semantic-complex guidance. Therefore, it is crucial to build a suitable dataset for training - in which the videos need to come from unstructured, complex real-world scenes and be consistent over a long duration. We finally decided to choose the autonomous driving scenarios to build such a dataset, with the goal of building a long video generation model that can simulate and navigate the unstructured complexity of real autonomous driving scenarios.
In this work, We construct a dataset namely FF-Drive1 with autonomous driving videos from 8 open-source datasets: BDD-Nexar~\cite{yu2020bdd100k}, nuScenes~\cite{caesar2020nuscenes}, vkitti2~\cite{cabon2020virtual}, 
Argoverse 2~\cite{wilson2023argoverse}, Brain4Car~\cite{jain2016brain4cars}, ApolloScape~\cite{Huang_2020}, Kitti-MOT~\cite{voigtlaender2019mots}, HOD~\cite{ha2023hod}.

\subsubsection{Construction Process}
For each video, we ask Video LLaVA through multiple rounds of dialogue in advance about the weather conditions, road background, vehicles ahead, congestion, zebra crossings, road arrows, traffic lights and their changes, as well as the vehicle's actions when encountering corresponding signs. Then, we organize the above dialogue and let Video LLaVA comprehensively summarize the video content, which are then be used as the final captions. Finally, the FF-Drive1 dataset contains 20k video clips with paired text annotations in detail, with the length of over 20 seconds and the frame rate of 24fps each video clip, and the total duration is about 112 hours. Considering social impact and privacy issues, we carefully remove driver face data from the dataset and manually filter the generated text captions to ensure legal compliance. The amount of data we collect is limited by time and resources, and in the future we plan to collect more long video data for training.

\begin{figure*}[t!]
  \centering
  \includegraphics[width=\textwidth]{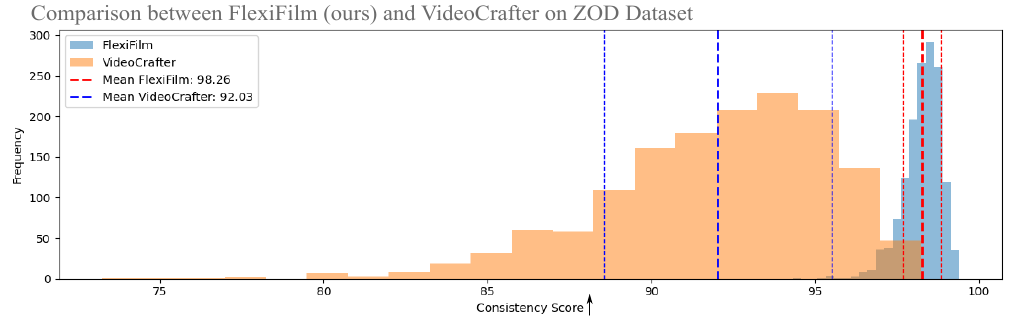}
  \caption{Consistency score comparison. The histogram shows significant improvements in inter-frame generative video consistency of our FlexiFilm compared to VideoCrafter (baseline).}
  \label{fig:cons}
\end{figure*}

\begin{figure}[t!]
  \centering
  \includegraphics[width=.48\textwidth]{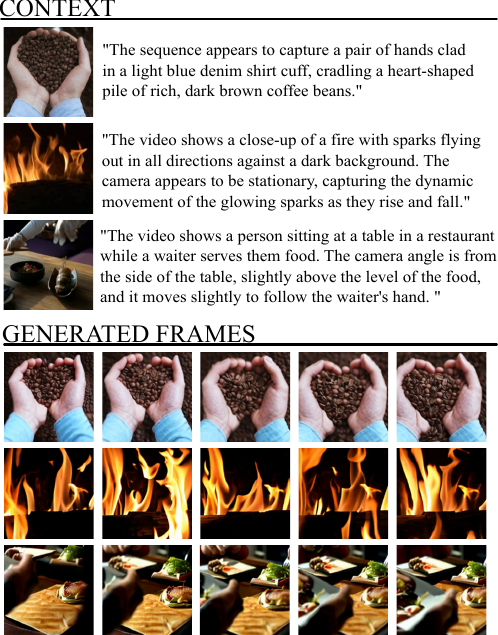}
  \caption{Natural-distributed video generation. The finetuned version of FlexiFilm can be generalize to video generation in natural distribution, which is consistent between generated frames and faithful to the conditional contexts.}
  \label{fig:n_exp}
\end{figure}

\section{Experiments}
\label{sec:exp}

\subsection{Experiment Settings}
\label{sec:exp-setting}
As discussed in Sec~\ref{sec:intro}, our FlexiFilm aims to generate long-and-consistent videos with multi-modal conditions: flexible visual conditions and long text descriptions. 
We qualitatively compare video generation results of FlexiFilm and baselines (VideoCrafter-1, Stable Video Diffusion) in Sec~\ref{sec:exp-baselines} with images and texts as conditions. And then we quantitatively compared the visual quality and structure consistency indicators of FlexiFilm and baselines in Sec~\ref{sec:exp-indicators}.

\subsection{Video Generation}
\label{sec:exp-baselines}
\subsubsection{Comparison with Baselines}
\label{sec:exp-short}
For video generation tasks, the baselines are not conditioned on multiple frames, so we use image-to-video version of FlexiFilm by setting the number of conditional frames to 1, and then conduct comparison experiments. As shown in Fig~\ref{fig:exp}, FlexiFilm can generate long video clips of the correct temporal and semantics relationship with the reference image, which lasts for 200+ frames. In contrast, VideoCrafter-1 and SVD can only generates short video clips (32- frames) in style of the reference image and lack temporal consistency between frames.

\subsubsection{Long and Short Video Generation in Different Domains}
\label{sec:exp-long}
For long and short video generation we train the FlexiFilm model on our 20K auto-driving dataset FF-Drive1 because long video clips with paired long text description is hard to get. In Fig~\ref{fig:exp}, we show the 224-frame video generation results inferred by the FlexiFilm in auto-driving scene with visual and text conditions, which is faithful to the detailed text description and the reference image. To show our method can be generalize to other domains, we finetune the FlexiFilm on the Pexels-1k~\cite{pexels_dataset} dataset, which contains 1000 short video clips of daily life and paired descriptions. In Fig~\ref{fig:n_exp}, we show the consistent and faithful video generation in natural distributions, which is inferred by the finetuned FlexiFilm. 

\begin{figure*}[p!]
  \centering
  \includegraphics[width=0.9\textwidth]{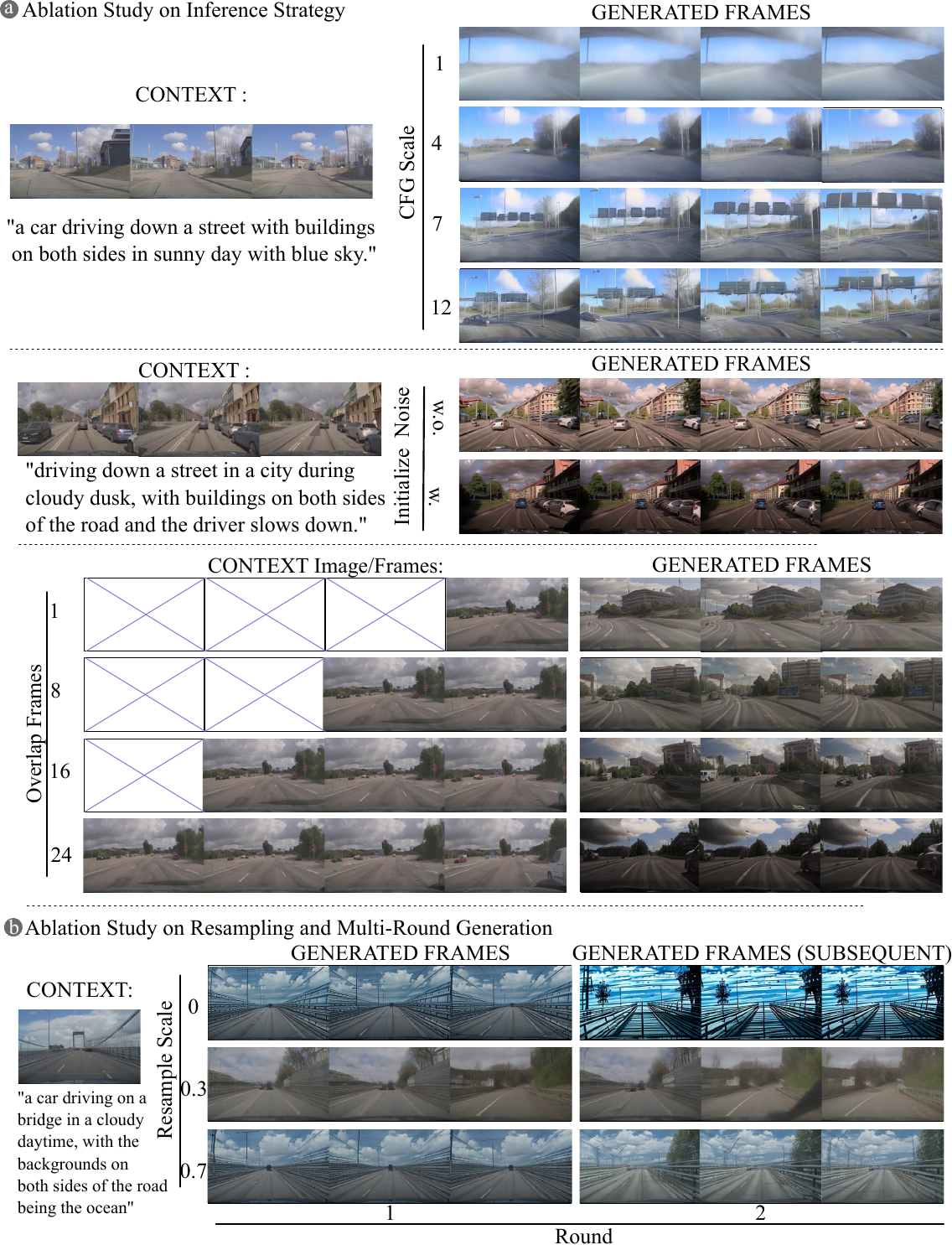}
  \caption{Ablation study of inference. In subplot (a), we first do an ablation study on the value of CFG scale to proved that the performance of the video generation model is indeed sensitive to it. Then, we do ablation study on our overlapping strategy, which improves the structural consistency between the generated object and the reference image. In subplot (b), we demonstrate that our resampling strategy helps alleviate the non-zero SNR problem under multiple rounds of inference.}
  \label{fig:ablation_eval}
\end{figure*}

\begin{figure*}[t!]
  \centering
  \includegraphics[width=0.95\textwidth]{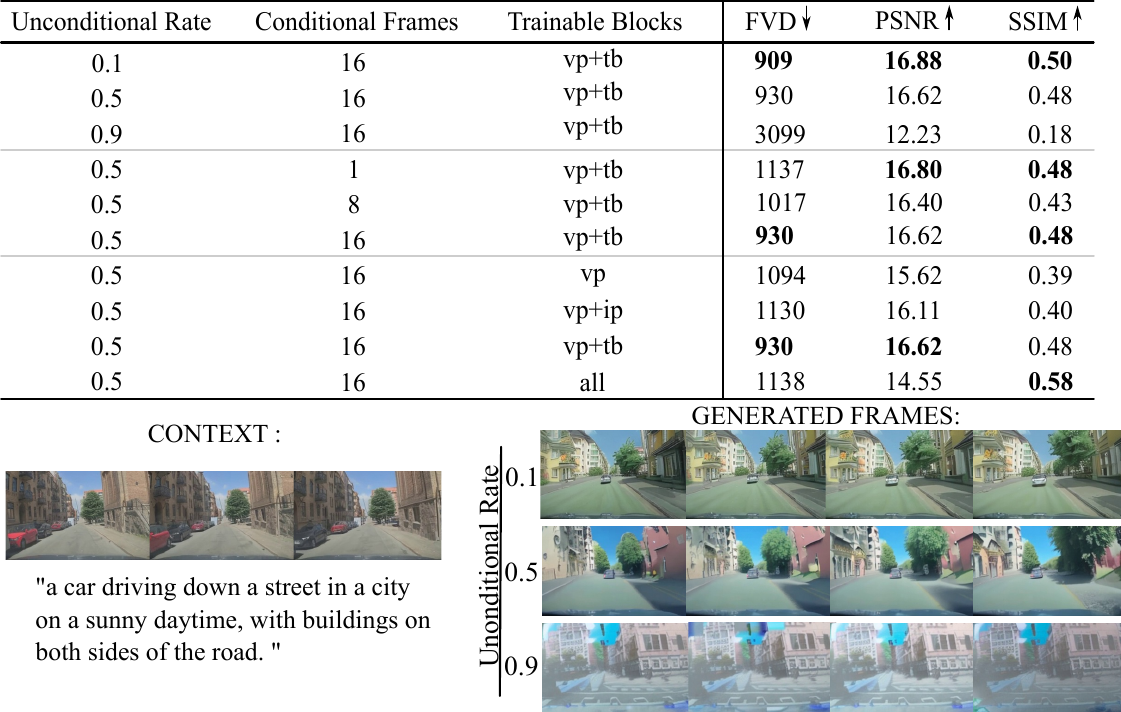}
  \caption{Comparison in consistency metric. The histogram upside shows significant improvements in generation consistency of FlexiFilm (ours) compared to VideoCrafter (baseline).}
  \label{fig:ablation_train}
\end{figure*}

\subsection{Quantitative Evaluation}
\label{sec:exp-indicators}
\subsubsection{Visual Quality}

For the quality of video generation, we calculate the FVD score~\cite{unterthiner2019accurate} in zero-shot video generation on UCF101~\cite{alibeigi2023zenseact} dataset. In Tab~\ref{tab:fvd_scores} we test FlexiFilm’s zero-shot capability on UCF101 and compared the FVD metric with existing mainstream video generation work, with the score of FlexiFilm are calculated by ourselves and the other model scores are excerpted from the corresponding papers. Note that different papers have different settings for this FVD score calculation, which will be explained in detail in the appendix. In our calculation for the FlexiFilm model, we set the generated video resolution to 256x256 and the number of frames to 16, with the class label and a single image as conditions.
The results in Tab~\ref{tab:fvd_scores} demonstrate that our method has better video quality in terms of FVD scores than the baseline.

\subsubsection{Consistency Indicator}
We define the consistency of generated videos as the coherence between frames. Specifically, we quantify the consistency score of each generated video as the average of the Clip Score~\cite{hessel2022clipscore} of every two adjacent frames. It is notable that the higher the consistency score, the more consistent the generated video is. As shown in Fig~\ref{fig:cons}(b), we compare the consistency score of FlexiFilm and VideoCrafter-1, demonstrating that our method outperforms the baseline model in inter-frame consistency.

\subsubsection{Other Indicators}
For other indicators like PSNR and SSIM ~\cite{unterthiner2019accurate, zhang2018unreasonable} on training dataset, we randomly sample a subset of size 1k from the training set for measurement. The measurement results of the models obtained by different training methods are given Fig~\ref{fig:ablation_train}.

\section{Ablation Study}

\subsection{Ablation Study on Co-Training}
During the training phase, we find that unconditional rate, conditional frames, and trainable blocks play the key roles, so we conducted ablation studies on these factors. First, we set the different unconditional probabilities and different conditional frames for co-training and train 6 models respectively. For different models obtained by training, we measured their performance scores on the training set respectively. Results in \ref{fig:ablation_train} show that the best setting is at unconditional rate 0.1.

\subsection{Ablation Study on Temporal Blocks}
We divide the model architecture into U-Net temporal blocks $tb$, video-projector spatial blocks $ip$ and video-projector temporal blocks $vp$, and train the model with full-parameter or part-of-parameter respectively, and find training U-Net temporal blocks together with video-projector temporal blocks has the best visual quality in FVD, PSNR and SSIM scores. As Fig~\ref{fig:ablation_train} shows, the best results for video generation are at only-temporal blocks training.

\subsection{Ablation Study on Inference Strategy}
In the inference stage, we find that the CFG scale and whether to initialize the noise of overlapping frames greatly affect the generation quality, and we conduct ablation studies on these factors. Experiment shows the generation quality is sensitive to different CFG scales as shown in Fig~\ref{fig:ablation_eval}(a). Due to the non-zero SNR problem, the generated results can be affected by the shape of the initial noise. To understand its potential effects, we conduct ablation studies to compare the inference results with initialized overlapping frame noises and non-initialized overlapping frame noises. In addition, we also compared the impact of different numbers of overlapping frames on the results during inference. The results in Fig~\ref{fig:ablation_eval}(a) shows that adding noise to the overlapping frames as the initial noise of the corresponding frames encourages structural faithful results.

\subsection{Ablation Study on Resampling Strategy}
Considering that we add a resampling process in the inference stage for solving the SNR problem and enabling multi-round generation, we conduct ablation studies on the subsequent resampling process.
For the resampling process, we compare the effects of different resampling scales for multiple inference rounds. The experimental results in Fig~\ref{fig:ablation_eval}(b) shows the resampling strategy alleviates color overexposure and helps to maintain better consistency during multi-round inference.

\section{Conclusion}
We present FlexiFilm, a latent video diffusion model tailored for long and consistent video generation. Aiming to solve the main failure modes of temporal inconsistency and overexposure in existing diffusion-based video generation, we design a delicate temporal conditioner to handle the intricate dynamics of long-duration videos, a co-training method to encourage temporal consistency and a resampling strategy to re-calibrate the noise levels during the generation process. Through comprehensive empirical validation, FlexiFilm has demonstrated superior performance over existing methods in both qualitative and quantitative metrics.



\bibliographystyle{ACM-Reference-Format}
\bibliography{sample-base}

\end{document}